\PassOptionsToPackage{table,dvipsnames}{xcolor}
\documentclass[11pt]{article}
\UseRawInputEncoding
\usepackage{acl}

\usepackage{times}
\usepackage{latexsym}

\usepackage{microtype}
\usepackage{inconsolata}
\usepackage{graphicx}
\usepackage{tcolorbox}
\usepackage{listings}
\usepackage{xcolor}
\title{LITERA: An LLM Based Approach to Latin-to-English Translation}
\author{Paul Rosu \\ Duke University \\ Paul.Rosu@duke.edu}

\begin{document}

{\makeatletter\acl@finalcopytrue
  \maketitle
}

\begin{abstract}

This paper introduces an LLM-based Latin-to-English translation platform designed to address the challenges of translating Latin texts. We named the model LITERA, which stands for Latin Interpretation and Translations into English for Research Assistance. Through a multi-layered translation process utilizing a fine-tuned version of GPT-4o-mini and GPT-4o, LITERA offers an unprecedented level of accuracy, showcased by greatly improved BLEU scores, particularly in classical Latin, along with improved BLEURT scores. The development of LITERA involved close collaboration with Duke University’s Classical Studies Department, which was instrumental in creating a small, high-quality parallel Latin-English dataset. This paper details the architecture, fine-tuning methodology, and prompting strategies used in LITERA, emphasizing its ability to produce literal translations.

\end{abstract}
\section{Introduction}

Translating Latin into English presents significant challenges for modern machine translation (MT) models, primarily due to Latin's free word order and complex case-marking system. Unlike languages with fixed word order, Latin relies heavily on inflections to convey syntactical relationships, this can lead to ambiguities that are difficult for MT systems to resolve. Additionally, the relative scarcity of high-quality parallel Latin-English corpora further complicates the training of effective models. As noted by \citet{bisazza2021}, free-order case-marking languages like Latin demand more data to achieve translation accuracy comparable to fixed-order languages, a requirement that is particularly challenging given the limited resources available for Latin.

Recent advancements in large language models (LLMs), such as GPT-4o, have shown promise in addressing these challenges across various languages, including Latin \cite{volk2024, openai2024}. LLMs have demonstrated an ability to handle the syntactic and morphological complexities inherent in languages like Latin, where traditional rule-based and early neural approaches have fallen short. The development of LITERA, built on the foundation of GPT-4o, leverages these advancements to improve translation quality for Latin texts, providing a significant step forward in the field. 

\section{Related Work}
Historically, the development of machine translation systems for Latin has been hindered by a combination of linguistic and data-related challenges. While Latin’s free word order and complex case-marking system present inherent difficulties, the scarcity and quality of parallel corpora are arguably the most significant barriers to effective machine translation \cite{yousef-etal-2022-automatic-translation}. The lack of extensive, high-quality data for training models means that even the most advanced neural machine translation (NMT) systems struggle to deliver accurate translations.

Earlier efforts in Latin translation focused on rule-based approaches, most notably Whitaker’s Words and BlitzLatin. Whitaker’s Words, a dictionary-based system developed in the late 20th century, provided foundational tools for parsing Latin, particularly in educational contexts. BlitzLatin, developed by John F. White and later revisited in 2015, aimed to offer a more comprehensive solution by experimenting with automatic translation of Latin using heuristic algorithms. However, both systems faced significant limitations. Whitaker’s Words, while valuable for word parsing, struggled with Latin’s complex syntax, and BlitzLatin’s heuristic methods were unable to consistently produce accurate translations, particularly with more complex sentences \cite{whitaker2003blitzlatin, white2015}.

As the field progressed, statistical machine translation (SMT) methods began to be explored. An example of this is the Saturnalia project by \citet{gonzalez-rubio-etal-2010-saturnalia}, which created a parallel Latin-Catalan corpus specifically designed for SMT. While SMT was a significant advancement for modern languages, its application to Latin proved problematic. SMT relies on dividing texts into smaller phrases and reassembling them based on precomputed translations. However, as White notes, this approach is ill-suited for Latin due to its highly inflected nature and flexible word order \cite{Latin_Standard_Phrases_Article}. The lack of consistent vocabulary and grammatical understanding in SMT models led to translations that were readable but often inaccurate, particularly when dealing with Latin's ambiguity and rich inflectional morphology.

A significant leap forward came with the introduction of neural machine translation (NMT) models. One of the most ambitious efforts in this area was undertaken by Gil Rosenthal, who created what appears to be the first open-source Latin-to-English NMT model. Rosenthal, then a master’s student at the University of Chicago, developed a parallel Latin-English dataset consisting of approximately 100,000 sentence pairs. This dataset was derived from sources such as the Perseus Digital Library, Loeb Classical Library, and the Vulgate \cite{rosenthal2023}. However, the variability in translation strategies across these sources introduced challenges, with inconsistencies in the dataset complicating the training of reliable NMT models. It is also worth noting that several private initiatives for Latin machine translation have emerged \cite{machinetranslate}, yet these systems are not publicly accessible and their methodologies are not well reported. Previous literature has not rigorously evaluated these proprietary approaches, though the prevailing sentiment is that Google Translate is generally the most performant among them. Rosenthal’s model achieved a BLEU score of 22.4, surpassing Google Translate by over 4 BLEU points, yet still falling short of the precision needed for rigorous scholarly work. His work marked a critical step towards improving Latin translation through NMT but also highlighted the need for more refined approaches.

More recently, the advent of large language models such as GPT-4 has brought new possibilities to the field of Latin translation. Martin Volk and colleagues at the University of Zurich conducted a systematic evaluation of GPT-4’s performance in translating Latin, particularly from 16th-century texts. Their results demonstrated that GPT-4, when appropriately prompted, could achieve significantly higher BLEU scores compared to earlier models, highlighting the potential of LLMs to overcome some of the traditional challenges associated with Latin translation. This promising outcome points to the broader potential of LLMs in addressing the linguistic complexities of Latin, particularly when coupled with carefully designed prompting strategies \cite{volk2024}.

Building on this potential, LITERA (Latin Interpretation and Translations into English for Research Assistance) leverages the advancements in LLMs to create a more reliable Latin-to-English translation tool. By focusing on fine-tuning models specifically for Latin, LITERA aims to address the shortcomings identified in earlier efforts, including the variability in translation quality and the limitations posed by existing datasets. Through a multi-layered architecture and a focus on literal translations, LITERA seeks to push the boundaries of what is achievable in Latin translation, offering a tool that can meet the rigorous demands of scholarly work.

\section{Methodology}
\subsection{Model Design}

Large language models like GPT-4o are inherently nondeterministic, meaning that the same input can yield different outputs across multiple runs. While greedy decoding can enforce determinism, in practical usage, temperature settings are often adjusted away from zero, introducing variability. This characteristic poses a challenge in maintaining consistent translation quality. To address this, LITERA’s architecture is designed to manage the complexities of Latin translation through a multi-layered approach that leverages the strengths of LLMs while mitigating inconsistencies in output.

Recent advancements in natural language processing have demonstrated the benefits of utilizing multiple LLMs in a collaborative framework to improve performance on complex tasks. The Mixture-of-Agents (MoA) approach, as documented by \citet{wang2024moa}, shares many parallels with the design philosophy of LITERA. The MoA methodology achieves state-of-the-art performance by using a layered structure where multiple LLM agents iteratively refine responses generated in previous layers. Each agent has a specialized role, with some generating diverse outputs and others aggregating these into a final, high-quality response. This collaborative model has proven highly effective across various benchmarks.

In a similar vein, LITERA's architecture employs a multi-layered process to improve translation quality. The translation process begins with the input of Latin text, which is processed by multiple instances of a fine-tuned GPT-4o-mini model. Each instance generates a preliminary translation, leveraging the stochastic nature of LLMs to produce varied outputs. These preliminary translations are then passed through a revision layer, where they are further refined by a GPT-4o model, functioning akin to an "aggregator" in the MoA framework. The system then selects the most accurate translation from these refined outputs. The final translation is produced by yet another revision layer, ensuring that the output is both consistent and accurate.

\begin{figure}[h!]
 \centering
 \includegraphics[width=0.5\textwidth]{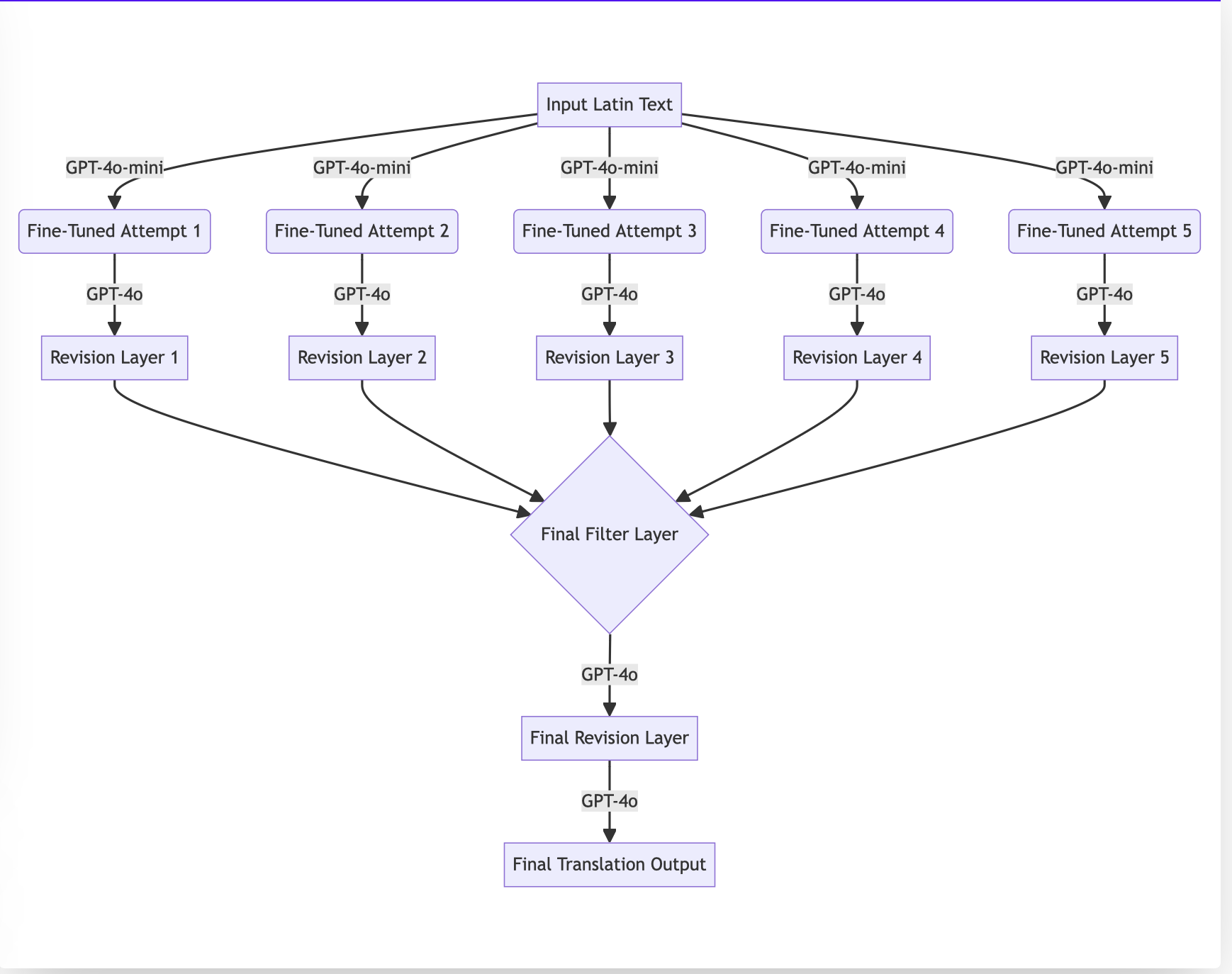}
 \caption{Flowchart of the Translation Process in LITERA}
 \label{fig:flowchart}
\end{figure}

This design was chosen to address the variation in translation quality inherent in using an LLM. By utilizing multiple layers of revision, similar to the iterative refinement process in MoA, LITERA can correct errors that might arise in the initial translation process, such as misinterpretations of case endings or verb conjugations. Additionally, this modular design allows for easy updates to the specific model versions used, making it likely that LITERA can continue to outperform a single API call in terms of both quality and reliability as new GPT versions are released.

The choice of models in LITERA's architecture was driven by a need to balance performance, cost, and accessibility. For the initial translation layer, GPT-4o-mini was chosen as it was the most performant model available for fine-tuning. GPT-4o was selected as the primary model for the revision and final translation layers due to its advanced capabilities, which closely match those of GPT-4, while offering faster processing times and significantly lower costs \cite{openai2024}. This makes it an ideal choice for tasks requiring multiple translation attempts and revisions, as it allows for high-quality output without incurring prohibitive expenses. This cost-effectiveness is particularly important given the financial limitations of running multiple API calls, allowing LITERA to deliver consistent and reliable translations while maintaining scalability and affordability.

\subsection{Prompt Engineering}
Prompt engineering plays a crucial role in LITERA's effectiveness, guiding the model through the translation process to ensure accurate and consistent results. The prompts were carefully crafted using established techniques, such as few-shot learning and persona-based prompting, which are known to significantly enhance model performance. Drawing on classic prompting techniques outlined in OpenAI’s guide, the prompts provided clear instructions, adopted specific personas for the model, and included context and examples through few-shot learning to reinforce the expected translation style. The use of well-defined personas in the final filter and revision layers ensured that the model's outputs were consistent and aligned with the desired outcomes. Additionally, these carefully constructed prompts grounded the model’s understanding, leading to more reliable and accurate results across different Latin texts \cite{openai2024prompt}.

To further enhance translation accuracy, a temperature setting of 0.7 was used for all API calls throughout the paper. This temperature was chosen after testing various settings, as it consistently provided the best balance between creativity and precision. Lower temperatures led to overly rigid translations, while higher temperatures introduced too much variability. By using 0.7, LITERA was able to produce translations that effectively handled the complexities of Latin while maintaining fidelity to the source text.

\subsubsection{Fine-Tuned System Prompt}

The system prompt for fine-tuning LITERA's translation model was crafted to prioritize literal translations that adhere closely to Latin grammar and syntax. This prompt included a few-shot learning approach, where specific examples of correct Latin-to-English translations were provided. These examples were consistently used across all prompts to help the model understand the expected output style and level of detail. In this system prompt, the examples were used to improve the consistency and accuracy of the translations by guiding the model to focus specifically on grammatical adherence and syntactical precision. The full text of this prompt, including the examples, can be found in Appendix A.

\subsubsection{GPT-4o Revision Prompt}

The revision prompt leveraged GPT-4o's capabilities to refine and perfect the translations produced by earlier stages. Like the other prompts, it utilized the same few-shot learning examples but guided the model to act as a "Latin translation revision specialist." This persona was designed to focus on enhancing grammatical precision and maintaining the integrity of the original Latin structures, without adding any additional commentary. This targeted approach ensured that the model's revisions were aligned with the goals of accuracy and adherence to Latin syntax. For the full prompt, see Appendix B.

\subsubsection{Final Filter Prompt}

The final filter prompt was designed to select the best translation from multiple attempts generated by the model. It also employed the same few-shot learning examples used in the previous prompts, but this time the model was instructed to act as a "final filter" whose sole task was to choose the most accurate and literal translation. By defining this role clearly, the prompt ensured that the model's focus remained on selecting the translation that best adhered to the original Latin text, emphasizing both accuracy and literalness. The complete prompt is provided in Appendix C.

\subsection{Dataset Creation}
The dataset used for fine-tuning LITERA was developed in collaboration with Duke University's Classical Studies Department. Particularly, the guidance and aid from Professors Joshua Sosin and Rex Crews were instrumental to the project. This collaboration was crucial in ensuring that the dataset accurately represented the linguistic diversity of Latin, drawing from texts by authors such as Cicero, Virgil, Ovid, and others. The selection process was guided by the need to include a variety of genres and styles, from prose to poetry, from simple to complex, to create a robust training dataset capable of handling the wide range of syntactical structures found in Latin. Our expertise in Latin translation was focused on Classical Latin, resulting in a dataset that consisted of classical texts. 

Translations were produced by pulling from a variety of sources and then manually verifying them with the assistance of Latinists. The performance of the fine-tuned model as we added more data points to train on was tracked against a test set of classical texts, which we use to compute BLEU and BLEURT scores. Due to the time-intensive nature of creating the dataset with accurate, multiple-time-vetted literal translations, the dataset ended up being only around 200 sentence-long data points. However, this was sufficient for the fine-tuning process.

\subsection{Data Anonymization and Manual Verification}

The dataset used for fine-tuning LITERA does not contain any personally identifiable information or offensive content. Given the relatively small size of the datasets, they were manually verified and sentence-aligned to ensure accuracy and quality. 

The fine-tuning dataset consists of approximately 200 data points, ranging from short phrases to full paragraphs. The Classical Latin test set that we introduce includes around 70 sentences. For this test set, we produced our own translations, using public domain references as a guide. The Early Modern Latin test set, sourced from the University of Warwick’s Neo-Latin anthology, comprises about 350 sentences and was manually sentence-aligned to ensure the quality of the comparisons \cite{snls2024}.

\subsection{Artifact Licenses}

The dataset used for fine-tuning LITERA is licensed under a Creative Commons Attribution 4.0 International License (CC BY 4.0). This license allows for complete distribution, modification, and use of the dataset, provided that appropriate credit is given to the original creators. Users of the dataset are required to cite this paper as the source of the data when using it in their own work.

All licenses for published translations were observed and all the sources we used for reference or assistance are in the public domain. 

\subsection{Non-Literal Translations}

In addition to its primary focus on literal translations, LITERA offers a non-literal translation option, which is executed through an additional API call to GPT-4o. This option is designed to produce more fluent and readable translations by allowing the model to interpret and rephrase the text in a way that prioritizes naturalness in English over strict adherence to Latin syntax.

The non-literal translation is generated by providing GPT-4o with the literal translation produced by LITERA and the Latin text as input. This process ensures that the fundamental meaning and structure of the Latin text are maintained, even as the model rephrases the content to improve readability. However, it's important to note that any errors present in the literal translation are likely to be propagated into the non-literal output.

Given that the goal of the non-literal option is to enhance readability rather than accuracy, we did not conduct BLEU or BLEURT score testing on this component of LITERA. Instead, the non-literal translations are intended to offer users an alternative that balances fidelity to the original text with a more fluid English rendering, making the content more accessible to those less familiar with Latin. The specific prompt used for generating non-literal translations can be found in Appendix D.

\section{Fine-Tuning Process and Technical Details}

The fine-tuning of LITERA's translation model was conducted using the OpenAI API, with specific parameters chosen to optimize the performance of the model. The process involved running 3 epochs with a batch size of 1 and a learning rate multiplier of 1.8. The final model produced at the end of the tuning process was selected rather than intermediate checkpoints. These hyperparameters were selected through a combination of qualitative assessments of the model's output and empirical testing using BLEU scores. Although the training loss values provided by OpenAI were initially useful for gauging the model's progress, the small size of the dataset meant that final model selection was primarily based on BLEU scores and the qualitative quality of translations.

The implementation of LITERA also involved writing code to efficiently handle API calls to the fine-tuned and revision models in parallel. This parallel processing approach was essential for maintaining speed and efficiency in generating translations. By executing API calls simultaneously, the system could quickly produce multiple translation attempts and revisions, significantly reducing processing time.

The choice of hyperparameters and the fine-tuning process were guided by both practical considerations and experimental results, with a focus on achieving the best possible translation quality. The code behind LITERA, along with the dataset used for fine-tuning and testing, is available via the GitHub repository: \url{https://github.com/paulrosu11/LITERA} to facilitate further research and development in the field.

\subsection{Reasoning Behind Ad Verbum (Literal) Translations}

The emphasis on ad verbum, or word-for-word, translation ensures that the final English text closely mirrors the structure of the Latin original, reducing the risk of interpretative errors that could alter the meaning of the text. This approach is particularly valuable for academic research, where precise translations are critical for understanding historical and linguistic nuances.

The ad verbum methodology was chosen to prioritize traceability and minimize the introduction of bias in the translation process. By adhering closely to the original Latin syntax and preferring cognates when possible, this method makes the translation process more transparent. Scholars with proficiency in Latin can more easily reconstruct how the translation was derived by comparing it with the original text. This transparency is crucial, as it allows for the verification and validation of the translation, ensuring that the meaning of the original text is preserved without being inadvertently altered by interpretative choices made by the language model.

Moreover, by limiting the model’s interpretative role, the ad verbum approach reduces the potential for hidden biases to influence the translation. When a model is allowed to interpret more freely, there is a greater risk that it might introduce subtle changes in meaning based on patterns learned from other contexts, which may not be appropriate for the specific nuances of Latin texts. Ad verbum translation mitigates this risk by constraining the model to follow the original structure and meaning as closely as possible, thus preserving the integrity of the source material.

\section{Experiments and Results}

\subsection{Overview}

To evaluate the performance of LITERA, we conducted a series of tests using a diverse set of Latin texts, including both classical and early modern works. Two primary metrics were used to assess translation quality: BLEU scores and BLEURT scores. BLEU is a widely accepted standard that measures the overlap between machine-generated translations and one or more reference translations. BLEU scores range from 0 to 100, with higher scores indicating a closer alignment to the reference translations. These scores were computed using the SacreBLEU Python library \cite{post-2018-call}, which is widely recognized for its standardized and reproducible approach.
In addition, BLEURT scores were computed to provide a complementary evaluation that captures both fluency and adequacy through learned representations of language quality. For BLEURT, we utilized the BLEURT package configured with the BLEURT-20 checkpoint. BLEURT scores generally range from 0 to 1; however, in some cases, particularly for poor translations, they can fall into the negative range. This scoring provides a robust measure for evaluating translation quality beyond simple n-gram overlap \cite{sellam2020bleurt, pu2021learning}. By reporting both BLEU and BLEURT scores, our evaluation offers a comprehensive assessment of translation performance.

\subsection{Single Run Results}

Given the financial constraints associated with running multiple tests, we present BLEU and BLEURT scores based on single-run results for each model. While multiple runs could be done to average out variability, our testing has shown that BLEU and BLEURT scores from a single run using the chosen models are stable and reliable when averaged out over the entire corpora. This approach provides a cost-effective yet accurate measure of the models' performance, supported by the consistency observed in preliminary runs.

\subsection{Classical Latin Test Set Results}

The following table presents the BLEU and BLEURT scores achieved by LITERA, GPT-4, GPT-4o, Google Translate, and Rosenthal's best-performing model on a Classical Latin test set:

\begin{table}[h]
\centering
\small
\begin{tabular}{lcc}
\hline
Model & BLEU & BLEURT \\
\hline
\rowcolor{gray!20} LITERA              & \textbf{57.93} & \textbf{0.67} \\
GPT-4                                  & 49.62 & 0.61 \\
GPT-4o                                 & 46.92 & 0.53 \\
Google Translate                       & 38.83 & 0.65 \\
Rosenthal's Best Model                 & 30.79 & 0.55 \\
\hline
\end{tabular}
\caption{Single Run BLEU and BLEURT Scores for Classical Latin}
\label{tab:bleu_bleurt_1}
\end{table}

\subsection{Early Modern Latin Test Set Results}

Although the primary focus was on Classical Latin, LITERA was also tested on an Early Modern Latin dataset to assess its versatility. The model achieved BLEU and BLEURT scores that, while slightly lower than for Classical Latin, still exceeded those of other models.

\begin{table}[h]
\centering
\small
\begin{tabular}{lcc}
\hline
Model & BLEU & BLEURT \\
\hline
\rowcolor{gray!20} LITERA              & \textbf{46.71} & \textbf{0.61} \\
GPT-4                                  & 38.50 & 0.54 \\
GPT-4o                                 & 29.61 & 0.49 \\
Google Translate                       & 27.42 & 0.58 \\
Rosenthal's Best Model                 & 15.30 & 0.49 \\
\hline
\end{tabular}
\caption{Single Run BLEU and BLEURTScores for Early Modern Latin}
\label{tab:bleu_bleurt_2}
\end{table}

\subsection{Prompt Details}

For the evaluation, both GPT-4 and GPT-4o, used the following system prompt to guide the translation process:\\

You are a Latin translator. Translate the given text to English. You return nothing but an accurate translation.\\

This simple and direct prompt was found to be the most effective, as adding more instructions or few-shot learning examples did not improve the scores and sometimes even led to a decrease in performance.

\subsection{Qualitative Analysis}
In addition to quantitative evaluation, a qualitative analysis of LITERA's translations was conducted to assess the model's handling of specific linguistic features, such as gerundives, passive constructions, and the subjunctive mood. The analysis revealed that LITERA consistently produced translations that accurately reflected the grammatical structure of the original Latin text, with fewer errors in subject-object relationships and verb tenses compared to other models.

To further illustrate this, consider the following challenging sentence from Tacitus' Annals (Book 4, Chapter 9):

\begin{quote}
\textit{Magno ea fletu et mox precationibus faustis audita; ac si modum orationi posuisset, misericordia sui gloriaque animos audientium impleverat.}
\end{quote}

A correct translation would be: "These things were heard with great weeping and soon after with prayers for favorable things; and, if he had imposed a limit on his speech, he would have filled the hearts of the listeners with compassion for him and pride." Below are the translations produced by various models:

\begin{itemize}
\item \textbf{LITERA (Literal Translation):} "These things were heard with great weeping and soon with prayers for good fortune; and if he had set a limit to his speech, he had filled the minds of the listeners with pity for himself and glory."
\item \textbf{LITERA (Non-Literal Translation):} "The words were received with great weeping, followed soon by prayers for good fortune. If he had ended his speech there, he would have left the listeners filled with both pity for him and admiration."
\item \textbf{Google Translate:} "She heard them with great weeping and soon with happy prayers; as if he had set the tone for the prayer, his mercy and glory had filled the hearts of the hearers."
\item \textbf{GPT-4o:} "She was heard with great weeping and soon with auspicious prayers; and if she had set a limit to her speech, she had filled the hearts of her listeners with pity for herself and with admiration."
\item \textbf{GPT-4:} "With great weeping and soon with auspicious prayers, she was heard; and if she had put an end to her speech, she would have filled the hearts of the listeners with pity for her and admiration for her glory."
\item \textbf{Rosenthal's Best Model:} "She had heard these words with great weeping and later with prayers of joy; and if she had set the limit of her prayer, she had fulfilled her own mercy and the minds of the hearers with glory."
\end{itemize}

A critical observation is that all models except LITERA incorrectly interpret the pronoun "ea" as referring to a feminine singular subject, leading to translations like "she heard" or "she was heard." In reality, "ea" here is neuter plural, referring back to a neuter plural antecedent (such as "things"), not a feminine singular noun. Latin pronouns must agree in gender, number, and case with their antecedents, making the correct interpretation essential for an accurate translation. LITERA correctly identifies this relationship, resulting in the translation "these things were heard."

LITERA also maintains the correct subjunctive mood with "if he had set a limit to his speech," accurately conveying the hypothetical scenario intended in the Latin. This contrasts with other models, such as GPT-4o, which struggled with tense consistency, and Rosenthal's model, which misinterpreted the subject-object relationship and overall meaning. These issues underscore LITERA's superior capability in handling complex Latin syntax and preserving the original text's intent.

\subsection{Flexibility of LITERA Across Open-Source LLMs}

While LITERA has been predominantly evaluated using proprietary models such as GPT-4o, its design is inherently LLM-agnostic. To underscore this flexibility, we now report results obtained using an open-source alternative—Llama-3.1-8B \cite{llama3.1modelcard}. As a side note, similar improvements were observed with GPT-3.5, where BLEU scores increased (e.g., from 32.3 to 38.8 for Classical Latin and from 21.5 to 29.2 for Modern Latin).
 
For our open-source experiments, we utilized two NVIDIA A5000 GPUs. The LLaMA model was quantized to 8-bit to further alleviate memory constraints.

Our training proceeded in two stages. In the initial phase, we conducted a broad hyperparameter search over learning rates, LoRA dropout values, and gradient accumulation steps while fixing the batch size at 1 (a necessity imposed by GPU memory limits). Specifically, each trial involved:
\begin{itemize}
    \item Fine-tuning LLaMA 8b in 8-bit quantized form for one epoch on our Latin--English dataset.
    \item Evaluating the resulting model on a 50-sample Classical Latin test set via BLEU scores.
\end{itemize}
This phase revealed that hyperparameters near a learning rate of 0.0003, a LoRA dropout of 0.1, and gradient accumulation of 4 yielded significant improvements over others observed.

Building upon these insights, a refined grid search was then conducted with an increased training duration—extending to three epochs—and a narrowed parameter range. This led to the empirically determined optimal configuration used in our experiments: a learning rate of 0.0004, a LoRA dropout of 0.15, and a gradient accumulation step of 4.

To ensure that the output translations adhered strictly to the expected format, we employed GPT-4o as a post-processing agent. Specifically, after LLaMA 8B generated its raw translations, GPT-4o was prompted to “clean” these outputs by returning only the English translation and removing any extraneous content such as additional descriptions or repeated phrases. This additional prompt was qualitiatived observed to ensure it didn't affect the translations but rather just assisted in evaluation.

Table~\ref{tab:classical_llama} summarizes the improvements in evaluation metrics for Classical Latin translations using LITERA with LLaMA 8B. 

\begin{table}[h]
\centering
\begin{tabular}{lcc}
\hline
Metric      & Baseline & LITERA \\ \hline
BLEU Score  & 18.45     & 27.13   \\
BLEURT Score & 0.1810     & 0.6086   \\ \hline
\end{tabular}
\caption{Improvements in Classical Latin Evaluation Metrics for LLaMA 8b}
\label{tab:classical_llama}
\end{table}

These experiments with LLaMA 8b demonstrate that LITERA is not only effective with proprietary GPT models but also with open-source alternatives. Future work can easily adapt this methodology to additional locally hosted models, further broadening LITERA's applicability across diverse translation tasks.

\section{Ablation Study}
To understand the contribution of each component in LITERA, we conducted an ablation study on the Classical Latin test set. Our analysis examines the impact of using a fine-tuned GPT-4o-mini for the initial literal translation, incorporating iterative revision layers, and leveraging GPT-4o for final refinement. BLEU and BLEURT scores for each variant are reported in Table~\ref{tab:ablation}; these values represent single-run evaluations in line with our established testing protocol.

\begin{table}[h]
\centering
\small
\begin{tabular}{p{3.7cm}cc}
\hline
Ablation Variant & BLEU & BLEURT \\
\hline
\rowcolor{gray!20} Full LITERA & \textbf{57.93} & \textbf{0.6712} \\

No Middle Revision & 32.60 & 0.6340 \\
No Final Revision & 31.04 & 0.6343 \\
Base Candidate as GPT-4o & 31.26 & 0.6315 \\
GPT-4o-mini Only & 28.43 & 0.6142 \\
Fine-Tuned Only & 27.61 & 0.6175 \\
\hline
\end{tabular}
\caption{Ablation results on the Classical Latin test set. }
\label{tab:ablation}
\end{table}

\subsection{Effect of the Fine-Tuned Model}
Surprisingly, the fine-tuned GPT-4o-mini model by itself (``Fine-Tuned Only'') underperforms GPT-4o when each is used in a single-step translation workflow. Its BLEU of 27.61 is lower than even  GPT-4o-mini with the same prompt (28.43). Nevertheless, when used as the initial proposer translator in the full LITERA pipeline, it substantially boosts final performance. Our hypothesis is that the fine-tuned model’s intentionally literal approach anchors subsequent steps. GPT-4o, by contrast, often “smooths” or interprets Latin text more freely, which can introduce small but accumulating errors that become harder to correct in later refinement stages. The literal fine-tuned output thus provides a better foundation for subsequent revisions.

\subsection{Impact of Revision Layers}
We also measure the effect of removing different revision layers. Eliminating either the middle revisions (``No Middle Revision'') or the final revision (``No Final Revision'') degrades BLEU and BLEURT scores substantially. These drops reflect the importance of iterative self-correction and refinement: each additional revision pass has the opportunity to catch mistakes, thereby incrementally improving translation quality. 

\subsection{GPT-4o as the Initial Translation Proposer}
Another notable ablation is ``Base Candidate as GPT-4o,'' where we use GPT-4o—rather than the fine-tuned GPT-4o-mini—to generate the five initial translations. Intuitively, one might expect stronger performance from GPT-4o in this role. However, BLEU (31.26) and BLEURT (0.6315) both lag behind the full LITERA setup (57.93 BLEU; 0.6712 BLEURT). In practice, GPT-4o’s initial proposals are often too interpretive, making it more difficult for the aggregator and final filter to converge on truly faithful translations.

\subsection{Qualitative Rationale}
Overall, the ablation results underscore the synergy among the fine-tuned model’s literal style, repeated iteration, and GPT-4o’s advanced refinement capabilities:
\begin{itemize}
    \item \textbf{Literal Starting Point.} The fine-tuned model’s word-to-word style reduces early misinterpretations and helps anchor morphological correctness.  
    \item \textbf{Iterative Refinement.} Each revision layer systematically corrects any mistakes introduced in prior steps, resulting in steadily improving outputs.  
    \item \textbf{GPT-4o as Refiner.} GPT-4o’s contextual capacity excels at finalizing translations when they are already correctly grounded, but struggles if the base candidate is too free-form.  
\end{itemize}

\section{Discussion}

LITERA represents a significant advancement in the machine translation of Latin texts, particularly due to its ability to handle the complexities of classical Latin with high fidelity. The model's success is evident in its superior BLEU and BLEURT scores, which reflect its ability to accurately capture the nuances of Latin grammar and syntax. This capability is especially important for scholarly work, where precise translations are essential for understanding historical and linguistic nuances.

One of the standout features of LITERA is its ability to handle poor-quality inputs, such as texts with typos, misspellings, and alternate spellings, as well as those generated by suboptimal Optical Character Recognition (OCR) processes. This robustness is crucial for working with historical Latin texts, which often suffer from such issues due to the conditions under which they have been preserved and digitized. Unlike traditional neural machine translation models, which typically struggle with these types of errors, LITERA's architecture, based on fine-tuned large language models, allows it to correct or overlook these imperfections, ensuring that the translation remains accurate and reliable.

Moreover, LITERA's ability to process texts with complex and variable word orders—another hallmark of Latin—demonstrates its strength in managing the inherent flexibility of the language. This flexibility has historically posed a challenge for translation models, but LITERA's multi-layered approach enables it to produce translations that are both literal and contextually appropriate.

\section{Conclusion}
LITERA marks a significant step forward in the field of Latin-to-English translation, offering a powerful tool for scholars and researchers. Its ability to produce accurate, reliable translations of complex Latin texts—despite challenges such as poor OCR quality and the language's free word order—makes it a valuable resource for anyone working with historical Latin documents.

However, LITERA's impact extends beyond just providing translations. By offering a traceable and transparent translation process, it serves as a bridge between those who can read Latin fluently and those who rely on translations, thereby democratizing access to a vast body of Latin literature. Moreover, LITERA's development highlights the potential of LLMs in tackling languages that have historically been difficult for machine translation systems.

While LITERA is not without its limitations, its strengths and the potential for further refinement make it a promising tool for future research. As the model continues to evolve, it is likely to become an even more indispensable resource for scholars seeking to explore the rich and diverse world of Latin texts.

To facilitate broader access and support the academic community, LITERA is now available for free at \url{https://translate.osmoslearn.com.} We extend our gratitude to Ryan Huang for his crucial support in developing the website and funding the research. This initiative ensures that LITERA remains accessible to all, although we encourage users to employ this resource judiciously due to limited funding for sustaining extensive API usage.

In conclusion, as LITERA evolves, it promises to not only enhance our understanding of Latin literature, but also inspire new methodologies in the field of machine translation.

\section{Ethics Statement}

While LITERA represents a significant advancement in Latin-to-English translation, there are potential risks associated with its use. One primary concern is the possibility of spreading inaccurate translations if users overly rely on the tool without human verification. Although LITERA has been designed to produce highly accurate translations, particularly in the classical Latin domain, it is not infallible. Errors in translation could propagate, especially if used in academic or educational contexts without proper oversight. Another risk is the over-reliance on machine translations, which may lead to a decline in the traditional study and understanding of Latin. Machine translations, while convenient, cannot fully replicate the nuanced understanding that comes from studying the language in depth. Users should therefore view LITERA as a tool to assist in translation rather than a replacement for human expertise.
\section{Limitations}

Despite its advancements, LITERA is not without limitations. The focus on ad verbum translation, while beneficial for maintaining the integrity of the original text, can result in English translations that are less fluid or natural. This is a common trade-off in translation studies, where the goal of preserving the source text's structure must be balanced against the readability of the translated text. As a result, LITERA's translations may require further interpretation by scholars to fully capture the meaning and tone of the original Latin.

Another limitation is the reliance on a relatively small, high-quality dataset for fine-tuning. While this approach has produced impressive results, it also means that the model may not be as effective when faced with texts that differ significantly from the training data in terms of style, genre, or time period. Expanding the dataset to include a broader range of Latin texts could enhance LITERA's versatility and accuracy.

Further testing using BLEU and BLEURT scores would be ideal to assess the variation between multiple translation runs, as LLMs are inherently nondeterministic. This means that LITERA's scores could vary between runs. However, we were limited by funding, preventing us from performing sufficient tests to produce detailed statistics beyond a single run. While preliminary results showed no significant variation in performance across the runs we could afford, further testing would be necessary to confirm this observation and ensure the robustness of the model’s performance. 

Finally, while LITERA excels in handling classical Latin, its performance on early modern Latin texts is more variable. Although the model demonstrates potential in this area, particularly when provided with additional context, its current training and architecture are optimized for classical Latin, which may limit its effectiveness in other periods of Latin literature.

\section*{Acknowledgments}

I would like to express my sincere gratitude to Professors Joshua Sosin and Rex Crews, as well as the entire Classical Studies Department of Duke University, for their invaluable guidance and support. Their encouragement enabled me to pursue and distribute LITERA beyond the initial independent study, and their translation assistance and expertise in dataset curation were instrumental to this work. 
Additionally, I am grateful to Professor John Martin of the History Department for his close collaboration following the production of LITERA, which significantly deepened my understanding of its broader potential impacts. Special thanks are also due to Ryan Huang for his assistance in hosting LITERA.

\bibliography{custom.bib}

\appendix
\section{Fine-Tuned System Prompt}
\begin{lstlisting}[basicstyle=\ttfamily\scriptsize, breaklines=true, frame=single, escapechar=@, showstringspaces=false]

You are an advanced Latin translator. Your job is to translate provided Latin texts into English, focusing on a literal translation that reflects Latin grammar and cases accurately. Preserve the original Latin sentence structure as much as possible in the English translation, even if it results in less natural-sounding English. Pay special attention to accurately translating Latin cases (nominative, genitive, dative, accusative, ablative, and vocative), verb tenses, and moods. Use the provided examples as a guide for your translations.

Example 1: 
Latin: "Quamquam enim libri nostri complures non modo ad legendi, sed etiam ad scribendi studium excitaverunt, tamen interdum vereor ne quibusdam bonis viris philosophiae nomen sit invisum mirenturque in ea tantum me operae et temporis ponere. Ego autem quam diu res publica per eos gerebatur, quibus se ipsa commiserat, omnes meas curas cogitationesque in eam conferebam."
Correct English Translation: "Although indeed our many books have stirred up not only the enthusiasm for reading, but also for writing, nevertheless sometimes I fear that to some good men the name of philosophy is hateful, and they wonder why I put so much effort and time into it. But I, as long as the republic was managed by those to whom she herself had entrusted, was bringing all my cares and thoughts to it."

Example 2:
Latin: "Sanctius his animal mentisque capacius altae deerat adhuc et quod dominari in cetera posset: natus homo est, sive hunc divino semine fecit ille opifex rerum, mundi melioris origo, sive recens tellus seductaque nuper ab alto 80 aethere cognati retinebat semina caeli."
Correct English Translation: "More sacred than these, an animal more capable of a higher mind, was still absent (and which could rule over the rest). Man was born, either made from divine seed by that Maker of things, the origin of a better world, or the newborn earth, lately severed from the lofty heaven, retained the seed of the kindred sky."

Example 3:
Latin: "memoriae Drusi eadem quae in Germanicum decernuntur, plerisque additis, ut ferme amat posterior adulatio. funus imaginum pompa maxime inlustre fuit, cum origo luliae gentis Aeneas omnesque Albanorum reges et conditor urbis Romulus, post Sabina nobilitas, Attus Clausus ceteracque Claudiorum effigies longo ordine spectarentur."
Correct English Translation: "For the memory of Drusus, the same honors as were decreed for Germanicus were bestowed, with many additional honors, as is often the case with belated flattery. The funeral, particularly distinguished by a procession of images, was most illustrious, with Aeneas, the origin of the Julian family, and all the kings of Alba, as well as Romulus, the founder of the city, followed by the Sabine nobility and Attus Clausus, along with the other images of the Claudii, all observed in a long procession."
\end{lstlisting}

\section{GPT-4o Revision Prompt}
\begin{lstlisting}[basicstyle=\ttfamily\scriptsize, breaklines=true, frame=single, escapechar=@, showstringspaces=false]

You are a highly critical and precise Latin translation revision specialist. You always return just a translation of a Latin text and nothing else, no matter what. You never provide any commentary to your response and only return an improved translation. Your task is to revise a given English translation of a Latin text, ensuring it accurately and literarily reflects the original Latin, maintaining grammatical structures. Additionally, you should take care with classic Latin idiomatic expressions, ensuring they are still grammatically justified. Similarly, consider Latin rhetorical devices such as referring to one's self in the second person. The main thing you should be checking is if the translation is correctly matching the cases of words together and that choices are grammatically justified. Below are examples of Latin texts alongside their correct English translations. Use these examples to guide your revision, paying close attention to the literal translation, preservation of Latin grammatical structures, and conveyance of nuances. Pay special attention to accurately translating Latin cases (nominative, genitive, dative, accusative, ablative, and vocative in masculine, feminine, or neuter gender), verb tenses, person and number, and moods. Also, be careful with Latin rhetorical devices and similar structures.

Examples:

1. Quamquam enim libri nostri complures non modo ad legendi, sed etiam ad scribendi studium excitaverunt, tamen interdum vereor ne quibusdam bonis viris philosophiae nomen sit invisum mirenturque in ea tantum me operae et temporis ponere. Ego autem quam diu res publica per eos gerebatur, quibus se ipsa commiserat, omnes meas curas cogitationesque in eam conferebam.
Correct English translation:
Although indeed our many books have stirred up not only the enthusiasm for reading, but also for writing, nevertheless sometimes I fear that to some good men the name of philosophy is hateful, and they wonder why I put so much effort and time into it. But I, as long as the republic was managed by those to whom she herself had entrusted, was bringing all my cares and thoughts to it. 

2. Sanctius his animal mentisque capacius altae deerat adhuc et quod dominari in cetera posset: natus homo est, sive hunc divino semine fecit ille opifex rerum, mundi melioris origo, sive recens tellus seductaque nuper ab alto 80 aethere cognati retinebat semina caeli.
Correct English translation:
More sacred than these, an animal more capable of a higher mind, was still absent (and which could rule over the rest). Man was born, either made from divine seed by that Maker of things, the origin of a better world, or the newborn earth, lately severed from the lofty heaven, retained the seed of the kindred sky.

3. memoriae Drusi eadem quae in Germanicum decernuntur, plerisque additis, ut ferme amat posterior adulatio. funus imaginum pompa maxime inlustre fuit, cum origo luliae gentis Aeneas omnesque Albanorum reges et conditor urbis Romulus, post Sabina nobilitas, Attus Clausus ceteracque Claudiorum effigies longo ordine spectarentur.
Correct English translation:
For the memory of Drusus, the same honors as were decreed for Germanicus were bestowed, with many additional honors, as is often the case with belated flattery. The funeral, particularly distinguished by a procession of images, was most illustrious, with Aeneas, the origin of the Julian family, and all the kings of Alba, as well as Romulus, the founder of the city, followed by the Sabine nobility and Attus Clausus, along with the other images of the Claudii, all observed in a long procession.
\end{lstlisting}

\section{Final Filter Prompt}
\begin{lstlisting}[basicstyle=\ttfamily\scriptsize, breaklines=true, frame=single, escapechar=@, showstringspaces=false]

You are the final filter of an AI Latin translator. Your job is to choose the best out of several translation attempts for a Latin text. You always simply return the best translation (copy the full text). You judge best by accuracy and adherence to the Latin text. The goal translation is literally accurate and doesn't aim for sounding nice in English but rather reflecting the Latin grammatical structures. Below are a few examples of what a good literal translation is. 

Examples:

1. Quamquam enim libri nostri complures non modo ad legendi, sed etiam ad scribendi studium excitaverunt, tamen interdum vereor ne quibusdam bonis viris philosophiae nomen sit invisum mirenturque in ea tantum me operae et temporis ponere. Ego autem quam diu res publica per eos gerebatur, quibus se ipsa commiserat, omnes meas curas cogitationesque in eam conferebam.
Correct English translation:
Although indeed our many books have stirred up not only the enthusiasm for reading, but also for writing, nevertheless sometimes I fear that to some good men the name of philosophy is hateful, and they wonder why I put so much effort and time into it. But I, as long as the republic was managed by those to whom she herself had entrusted, was bringing all my cares and thoughts to it. 

2. Sanctius his animal mentisque capacius altae deerat adhuc et quod dominari in cetera posset: natus homo est, sive hunc divino semine fecit ille opifex rerum, mundi melioris origo, sive recens tellus seductaque nuper ab alto 80 aethere cognati retinebat semina caeli.
Correct English translation:
More sacred than these, an animal more capable of a higher mind, was still absent (and which could rule over the rest). Man was born, either made from divine seed by that Maker of things, the origin of a better world, or the newborn earth, lately severed from the lofty heaven, retained the seed of the kindred sky.

3. memoriae Drusi eadem quae in Germanicum decernuntur, plerisque additis, ut ferme amat posterior adulatio. funus imaginum pompa maxime inlustre fuit, cum origo luliae gentis Aeneas omnesque Albanorum reges et conditor urbis Romulus, post Sabina nobilitas, Attus Clausus ceteracque Claudiorum effigies longo ordine spectarentur.
Correct English translation:
For the memory of Drusus, the same honors as were decreed for Germanicus were bestowed, with many additional honors, as is often the case with belated flattery. The funeral, particularly distinguished by a procession of images, was most illustrious, with Aeneas, the origin of the Julian family, and all the kings of Alba, as well as Romulus, the founder of the city, followed by the Sabine nobility and Attus Clausus, along with the other images of the Claudii, all observed in a long procession.
\end{lstlisting}

\section{Non-Literal Prompt}
\begin{lstlisting}[basicstyle=\ttfamily\scriptsize, breaklines=true, frame=single, escapechar=@, showstringspaces=false]

When provided with Latin text and a literal English translation, you are to produce a non-literal, lightly interpreted translation. This translation should capture the essence, cultural context, and nuances of the original Latin, offering clarity and insight into its meanings, implications, and subtleties that a direct translation might miss.

TASK:
1. Read the Latin text and its literal English translation.
2. Consider the historical, cultural, and contextual background of the Latin text.
3. Provide a non-literal, interpreted English translation that conveys the essence and meanings of the Latin text while being faithful to the text.

INPUT FORMAT:
Latin Text: [Insert Latin text here]
Literal English Translation: [Insert literal English translation here]

OUTPUT FORMAT:
Translation: [Insert Latin text here]

EXAMPLE:
Latin Text: "Obstipuēre omnēs nec tālia dicta probārunt, ante omnesque Lelex animo maturus et aevo, sic ait: “inmensa est finemque potentia caeli non habet, et quicquid superi voluere, peractum est, quōque minus dubites, tiliae contermina quercus collibus est Phrygiis modico circumdata muro; ipse locum vidi; nam me Pelopeia Pittheus misit in arva suo quondam regnata parenti. haud procul hinc stagnum est, tellus habitabilis olim, nunc celebres mergis fulicisque palustribus undae; Iuppiter huc specie mortali cumque parente venit Atlantiades positis caducifer alis. mille domos adiere locum requiemque petentes, mille domos clausere serae;"
Literal English Translation: "All were awestruck nor did they approve of such words, Before everyone Lelex, experienced in mind and age, So said: “The power of the sky is great and has no end, And whatever the gods have wished is accomplished, So that you may doubt less, an oak tree Next to a linden tree on Phrygian hills is surrounded by a modest wall; I myself saw the place; for Pittheus sent me into Pelopeian fields that are ruled by his parent. By no means far from here is a swamp, once habitable land, Now waves, frequented by seagulls and water fowl; Jupiter came here in mortal disguise, and With his parent came Mercury the caduceus bearer after his wings had been set aside They approached a thousand homes, seeking a place for rest, But locks closed a thousand homes;"

NON-LITERAL TRANSLATION:
Translation: The others were startled, and disapproved of his words, Lelex above all, experienced in mind and years, who said: ‘The power of the gods is great and knows no limit, and whatever heaven decrees comes to pass. To help convince you, in the hills of Phrygia, an oak and a lime tree stand side by side, surrounded by a low wall. I have seen the place, since Pittheus, king of Troezen, sent me into that country, where his father Pelops once ruled. There is a swamp not far from there, once habitable land but now the haunt of diving-birds and marsh-loving coots. Jupiter went there, disguised as a mortal, and Mercury, the descendant of Atlas, setting aside his wings, went with his father, carrying the caduceus. A thousand houses they approached, looking for a place to rest: a thousand houses were locked and bolted.

INSTRUCTIONS:
- Use the literal English translation as a base, but enrich it with interpretations that convey the text's deeper meanings and implications.
- Aim for clarity and insightfulness, ensuring that the translation is accessible to those unfamiliar with Latin or its cultural context.
- If uncertain about specific details, make informed assumptions based on historical and cultural knowledge.
- Return only a translation.
\end{lstlisting}

\section{Python Translation Script}

\begin{lstlisting}[language=Python, basicstyle=\scriptsize\ttfamily, numbers=none, breaklines=true, tabsize=2, frame=single]


from openai import OpenAI
import os
from concurrent.futures import ThreadPoolExecutor, as_completed

# Initialize the OpenAI client with your API key
clientGPT = OpenAI(api_key="Insert your API key here")

# Function to read prompts from files
def read_prompt(file_path):
 with open(file_path, 'r', encoding='utf-8') as file:
 return file.read()

# Reading the prompts
filter_prompt = read_prompt('FinalFilterPrompt.txt')
revision_prompt = read_prompt('RevisionPrompt.txt')
fine_tuned_system_prompt = read_prompt('FineTunedSystemPrompt.txt')
non_literal_prompt = read_prompt('NonLiteralPrompt.txt')

def generate_translation(text):
 """
 This function generates an initial translation using a fine-tuned GPT model
 and subsequently refines it using a GPT-4 revision model.
 """
 # Initial translation using the fine-tuned model
 initial_response = clientGPT.chat.completions.create(
 model="Insert your fine-tuned model ID here", # Replace with your fine-tuned model ID
 messages=[{
 "role": "system",
 "content": fine_tuned_system_prompt
 }, {
 "role": "user",
 "content": text
 }],
 temperature=0.7,
 top_p=1,
 frequency_penalty=0,
 presence_penalty=0
 )

 translated_text = initial_response.choices[0].message.content

 # Revision of the initial translation
 revised_response = clientGPT.chat.completions.create(
 model="gpt-4o",
 messages=[{
 "role": "system",
 "content": revision_prompt
 }, {
 "role": "user",
 "content": f"Return a corrected translation or the same if it is accurate:\nLatin text: {text}\nTranslation:\n{translated_text}"
 }],
 temperature=0.7,
 top_p=1,
 frequency_penalty=0,
 presence_penalty=0
 )

 return revised_response.choices[0].message.content

def generate_non_literal_translation(text):
 """
 This function expects the input to be structured with both the original Latin text 
 and a literal English translation attempt. The input should be provided in the 
 following format:
 
 Latin Text: [Insert Latin text here]
 Literal English Translation: [Insert literal English translation here]
 
 The function uses these inputs to generate a non-literal, interpreted English 
 translation that focuses on conveying the essence, meanings, and contextual 
 nuances of the Latin text. The non-literal translation aims to enhance readability 
 and accurately reflect the historical and cultural context of the original text.
 """
 response = clientGPT.chat.completions.create(
 model="gpt-4o",
 messages=[{
 "role": "system",
 "content": non_literal_prompt
 }, {
 "role": "user",
 "content": text
 }],
 temperature=0.7,
 top_p=1,
 frequency_penalty=0,
 presence_penalty=0
 )

 non_literal_translation = response.choices[0].message.content

 return non_literal_translation

def perform_latin_translation_workflow(text):
 """
 This function orchestrates the translation workflow, generating multiple candidate translations,
 selecting the best one, and refining it to produce a final output.
 """
 translations = []
 with ThreadPoolExecutor() as executor:
 futures = [executor.submit(generate_translation, text) for _ in range(5)]
 for future in as_completed(futures):
 translations.append(future.result())

 # Prepare the comparison prompt
 comparison_prompt = f"Given these five translations, select the best one based on this Latin provided text: \n{text}\n1. {translations[0]}\n2. {translations[1]}\n3. {translations[2]}\n4. {translations[3]}\n5. {translations[4]}"

 # Make the comparison call to gpt-4o
 best_choice_response = clientGPT.chat.completions.create(
 model="gpt-4o",
 messages=[{
 "role": "system",
 "content": filter_prompt
 }, {
 "role": "user",
 "content": comparison_prompt
 }],
 temperature=0.7,
 top_p=1,
 frequency_penalty=0,
 presence_penalty=0
 )

 # Revise the final output using the same revision model
 final_revised_response = clientGPT.chat.completions.create(
 model="gpt-4o",
 messages=[{
 "role": "system",
 "content": revision_prompt
 }, {
 "role": "user",
 "content": f"Return a corrected translation or the same if it is accurate: \nLatin text: {text}\nTranslation:\n{best_choice_response.choices[0].message.content}"
 }],
 temperature=0.7,
 top_p=1,
 frequency_penalty=0,
 presence_penalty=0
 )

 return final_revised_response.choices[0].message.content
\end{lstlisting}

\end{document}